\documentclass[sigconf, screen]{acmart}

\usepackage{multirow}
\usepackage{afterpage}
\usepackage[normalem]{ulem}
\usepackage{colortbl}
\usepackage{arydshln}
\usepackage{ragged2e}

\renewcommand\footnotetextcopyrightpermission[1]{}

\AtBeginDocument{%
  }

\setcopyright{acmlicensed}
\copyrightyear{2025}
\acmYear{2025}

\acmConference[arXiv]{}{2025}{}

\acmBooktitle{arXiv}

\begin{document}

\title{KASportsFormer: Kinematic Anatomy Enhanced Transformer for 3D Human Pose Estimation on Short Sports Scene Video}


\author{Zhuoer Yin}
\affiliation{
    \institution{Nagoya University}
    \city{Nagoya}
    \country{Japan}
}
\email{yin.zhuoer@g.sp.m.is.nagoya-u.ac.jp}

\author{Calvin Yeung}
\affiliation{
    \institution{Nagoya University}
    \city{Nagoya}
    \country{Japan}
}
\email{yeung.chikwong@g.sp.m.is.nagoya-u.ac.jp}

\author{Tomohiro Suzuki}
\affiliation{
    \institution{Nagoya University}
    \city{Nagoya}
    \country{Japan}
}
\email{suzuki.tomohiro@g.sp.m.is.nagoya-u.ac.jp}

\author{Ryota Tanaka}
\affiliation{
    \institution{Nagoya University}
    \city{Nagoya}
    \country{Japan}
}
\email{tanaka.ryota@g.sp.m.is.nagoya-u.ac.jp}

\author{Keisuke Fujii}
\affiliation{
    \institution{Nagoya University}
    \city{Nagoya}
    \country{Japan}
}
\email{fujii@i.nagoya-u.ac.jp}

\renewcommand{\shortauthors}{Yin et al.}



\begin{abstract}

Recent transformer based approaches have demonstrated impressive performance in solving real-world 3D human pose estimation problems. Albeit these approaches achieve fruitful results on benchmark datasets, they tend to fall short of sports scenarios where human movements are more complicated than daily life actions, as being hindered by motion blur, occlusions, and domain shifts. Moreover, due to the fact that critical motions in a sports game often finish in moments of time (e.g., shooting), the ability to focus on momentary actions is becoming a crucial factor in sports analysis, where current methods appear to struggle with instantaneous scenarios. To overcome these limitations, we introduce KASportsFormer, a novel transformer based 3D pose estimation framework for sports that incorporates a kinematic anatomy-informed feature representation and integration module. In which the inherent kinematic motion information is extracted with the Bone Extractor (BoneExt) and Limb Fuser (LimbFus) modules and encoded in a multimodal manner. This improved the capability of comprehending sports poses in short videos. We evaluate our method through two representative sports scene datasets: SportsPose and WorldPose. Experimental results show that our proposed method achieves state-of-the-art results with MPJPE errors of 58.0mm and 34.3mm, respectively. Our code and models are available at: \url{https://github.com/jw0r1n/KASportsFormer}.

\end{abstract}




\maketitle

\begin{figure}[htbp]
    \centering
    \includegraphics[width=1.0\linewidth]{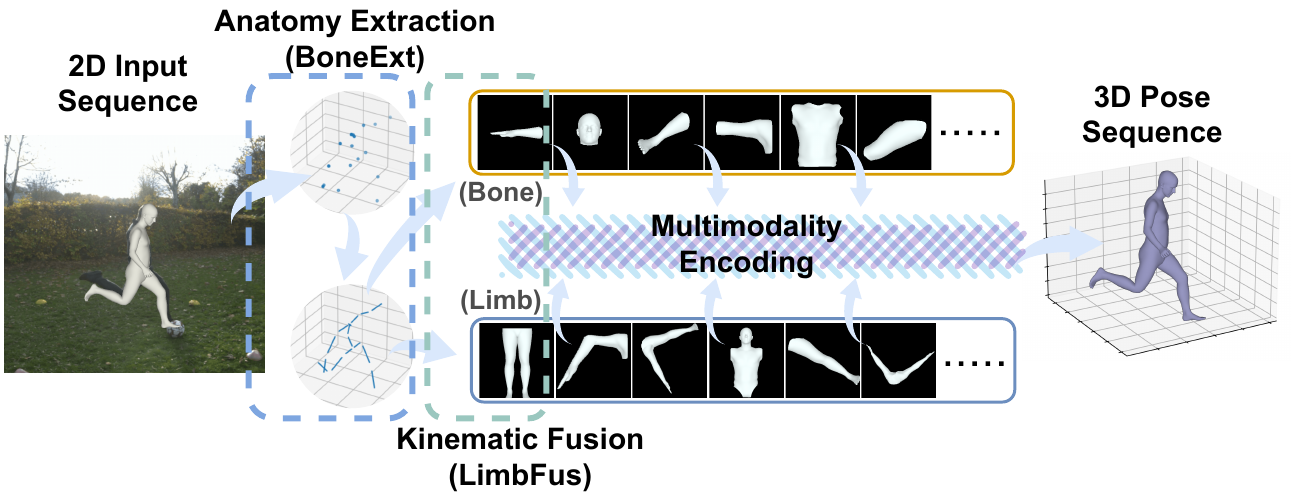}
    \caption{\textbf{Illustration of our proposed method.} Anatomy structures are extracted and composed into bones and limbs, which are interacted in a multimodality strategy.}
    \Description{The pipeline of our proposed KASportsFormer method.}
    \label{fig:overall}
\end{figure}

\section{Introduction}
\label{sec:introduction}

Human Pose Estimation (HPE) has always been a vital topic in computer vision. Unlike 2D HPE, 3D HPE aims at reconstructing human motion in 3D perspectives from monocular observations, thus being a more challenging machine learning task with significant applications from human-computer interactions \cite{gong2022meta, bauer2023weakly, 3dhumanposehci}, to virtual or augmented reality \cite{mehta2017vnect, handpose3d}, sports games \cite{yeung2024autosoccerpose, yeung2025athletepose3d}. Obtaining 3D poses directly from 2D observations has the tendency of posing low accuracy problems \cite{pavlakos2018ordinal}, which is an unignorable drawback in human motion analysis. Current 3D HPE works are mostly decomposed into two stages \cite{zhu2023motionbert,xu2024finepose}: 1) 2D Pose Estimation: detecting 2D human keypoints from image or video sources, and 2) 2D-3D lifting: mapping 2D human keypoints to 3D human poses.

Consider sports scenarios, by analyzing athletes' 3D pose sequences, the technical performance of their motion can be quantitatively measured and assessed \cite{tanaka2024,yeung2025athletepose3d}. Despite its importance, existing monocular 3D HPE works \cite{peng2024ktpformer, zhu2023motionbert} are commonly challenged by: 1) Action Generalizability: Although achieving robust and accurate results on daily life motions \cite{h36m_pami}, current methods are prone to losing performance in sports actions due to the intrinsic motion complexity \cite{ingwersen2023sportspose}. For example, more fast-moving actions and self-occlusions caused by spins. 2) Temporal reliance: Critical sports actions (e.g. shooting, passing) usually extend within a few frames, while existing approaches face a rapid accuracy decline when processing short videos since their parameters are fitted for video clips with a longer frame length (i.e., 243 frames).

One strategy to address these limitations is fine-tuning 3D HPE models with the extra domain-specific data as fine-tuned MotionAGFormer \cite{mehraban2024motionagformer} with a self-supervised method \cite{suzuki2024pseudo}. Nevertheless, this approach is still confronting dilemmas of visual contexts in sports videos, including background, lighting, and camera angle variations. 
Meanwhile, extending temporal information alone does not fully account for the detailed representation of sports movements. As demonstrated by Peng \textit{et al.} \cite{peng2024ktpformer}, modeling kinematic dependencies within the spatial dimension can still lead to substantial performance gains. Building on this insight, we assume that further potential lies in explicitly exploring human anatomical structures to better capture the underlying motion patterns.

In this work, we propose a Kinematic Anatomy enhanced Sports Transformer (\textbf{\textit{KASportsFormer}}) for 3D HPE. As shown is Figure~\ref{fig:overall}, our KASportsFormer is composed of a bone-wise anatomy structure extractor (\textbf{\textit{BoneExt}}), a limb combination fuser (\textbf{\textit{LimbFus}}), and a multimodality token mixing mechanism incorporated inside a spatio-temporal Transformer \cite{vaswani2017attention}. To be specific, the BoneExt extracts the original joint into an equivalent vector-like format, encoding both bone directions and lengths. The LimbFus collects and aggregates these separate bone parts to symbolize limbs not only in intuitive topological patterns but also in hypothetical \cite{chen2023hdformer} orders to enhance kinematic interactions inside movements. The contributions of our work can be summarized as follows:

\begin{itemize}
    \item We propose KASportsFormer, a novel method that models interactions between kinematic bone and limb tokens through a multimodal communication framework.
    \item Our approach effectively improves 3D HPE performance in sports scenarios, particularly under conditions with limited temporal information.
    \item Extensive experiments reveal that our KASportsFormer is able to obtain reconstruction result improvements on SportsPose and WorldPose datasets and achieves state-of-the-art.
\end{itemize}

\section{Related Work}
\label{sec:relatedwork}

\noindent \textbf{3D Human Pose Estimation.} Analogous to 2D HPE, 3D HPE aims to estimate human poses but in the 3D space. Current 3D HPE works can be categorized into either direct 3D estimation or 2D-3D lifting approaches. Direct estimation methods \cite{zhou2019hemlets,sun2018integral,pavlakos2018ordinal, zhang2021direct} perform 3D inference directly from 2D sources without any intermediate steps. With the maturing of 2D HPE, the 2D-3D lifting method is more commonly adopted in recent works, where 3D estimation is first performed with an off-the-shelf 2D pose detector \cite{xu2022vitpose,sun2019deep,chen2018cascaded} and then lifted into 3D space \cite{pavlakos2017coarse, martinez2017simple,  pavllo20193d,zhang2022mixste,zheng20213d,zhang2025posemagicefficienttemporally}. 3D HPE can likewise be categorized into monocular and multiview \cite{chun2023learnable,reddy2021tessetrack,zhang2021adafuse} based on input sources. Although the multi-view contains more motion contexts, setting specific multiple camera angles is challenging in sports applications. 3D HPE can also be summarized as single or multiple person-based on task targets, where multiperson approaches \cite{jiang2024worldpose, Mehta_2020, zhou2021hemletsposhlearningpartcentric, mehta2018singleshotmultiperson3dpose} are strict with computational resources, which remains challenging in sports scenarios. In this work, we follow the 2D-3D lifting methodology and utilize monocular single-person videos as the input. 

\noindent \textbf{Transformer-based 3D HPE.} Transformer \cite{vaswani2017attention} has been demonstrating powerful comprehension capability in encoding long-range context dependencies in NLP tasks. Recently, Transformer-based methods are showing promising results in tackling visual tasks \cite{su2023towards, yu2022coca, zong2023detrs}, including monocular 3D HPE problems\cite{tang20233d, liu2020attention, li2022exploiting, xu2024finepose, li2024hourglass, li2022mhformer, islam2024multi, shan2022p}. Concerning 3D HPE, based on the attempt of vanilla head \cite{li2022exploiting}, PoseFormer \cite{zheng20213d} made the first trial of the Transformer structure-based approach for 3D pose reconstruction, and PoseFormerV2 \cite{zhao2023poseformerv2} reduces its computational cost by introducing a time-frequency conversion block. MHFormer \cite{li2022mhformer} studies the multi-hypothesis caused by the intrinsic lack of depth. MixSTE \cite{zhang2022mixste} alternately concatenates spatial and temporal transformer blocks to track the joint trajectory over frame ranges, while STCFormer \cite{tang20233d} splits the data stream into spatial and temporal transformer components in parallel. HSTFormer \cite{qian2023hstformer} utilized a hierarchical module for better encoding of spatial and temporal relations. Recently, P-STMO \cite{shan2022p} proposed masked pose modeling and a self-supervised pertaining strategy. MotionBERT \cite{zhu2023motionbert} likewise proposed a pertaining stage to recover the underlying 3D motion. KTPFormer \cite{peng2024ktpformer} leverages motion topology and kinematic consistency, utilizing prior knowledge to enhance patched attention. MotionAGFormer \cite{mehraban2024motionagformer} attempts to effectively capture motion correlations both locally and globally through combining GCN in parallel with attention. In addition, GLA-GCN \cite{yu2023gla} globally models the 2D representations and locally recovers the 3D pose joints. D3DP \cite{shan2023diffusion} designs the diffusion denoiser for 3D HPE. FinePose \cite{xu2024finepose} draws prompt learning (e.g., CLIP) into denoising. PoseMagic \cite{zhang2025posemagicefficienttemporally} recovers human motion with Mamba structure for better temporal consistency. 

\noindent \textbf{Anatomy Feature Involved Methods.} In order for a superior understanding of the human skeleton, some work introduces human anatomy features into the reconstruction process \cite{chen2021anatomy, hsu2024blapose, chen2023hdformer}. Chen et al. \cite{chen2021anatomy} innovatively explores anatomy features in 3D HPE and demonstrates that the prediction of bone direction and bone length benefits motion reconstruction. BLAPose \cite{hsu2024blapose} proposes the bone length adjustment, which demonstrates effectiveness in fine-tuning existing models. Moreover, HDFormer \cite{chen2023hdformer} exploits not only joint-bone but also hyperbone-joint interactions, which is proven to bring improvements in inference efficiency. Despite taking advantage of bone prediction being capable of promoting results, the interacting manner of bone features has not been well considered, nor the potential kinematic motion pattern, both locally and globally. In our work, we demonstrate the significance of utilizing bone features to improve sports action understanding, emphasizing the advantage of exploring intrinsic kinematic motion characteristics.

\section{Method}
\label{sec:method}

\begin{figure*}[htbp]
    \centering
    \includegraphics[width=1\linewidth]{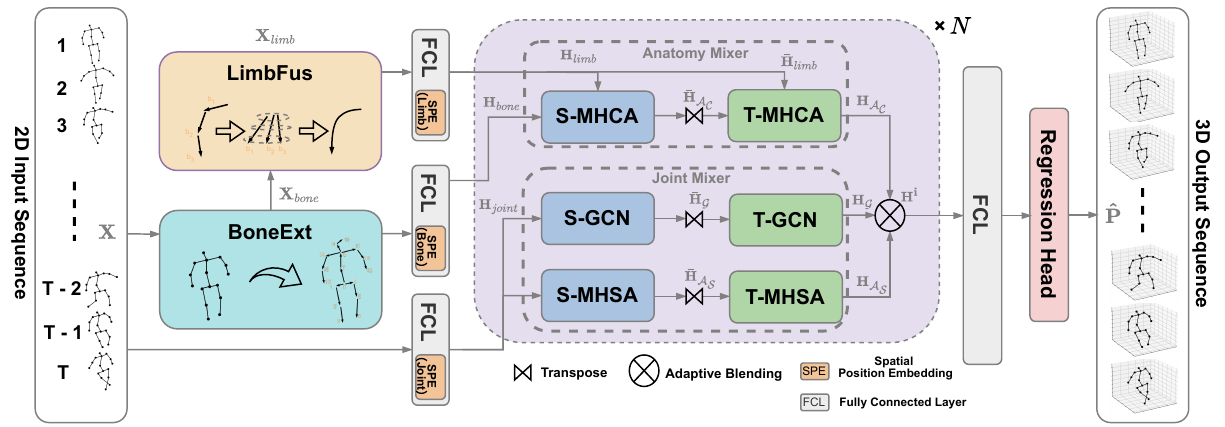}
    \caption{The architecture of our proposed Kinematic-Anatomy Sports Transformer (KASportsFormer). Anatomy Bone Extractor (BoneExt) constructs bone vectors from input $\mathbf{X}$. Kinematic Limb Fusion (LimbFus) aggregates separated bones into composed limbs. The mutual interactions are explored with a multimadality manner inside a Spatio-Temporal Transformer.}
    \Description{The detailed architecture of our proposed method.}
    \label{fig:fig-pipeline}
\end{figure*}

\subsection{Overall Pipeline}

Following the seq2seq 3D HPE pipeline as by Zhang \textit{et al.} \cite{zhang2022mixste}, the objective of our work is to lift the 2D monocular skeleton observation into corresponding 3D pose sequences. To accomplish this envision, we propose the KASportsFormer structure, which utilizes anatomical bone features with a multiple stream transformer to map pose sequences. An overview of our method is shown in Figure~\ref{fig:fig-pipeline}. 

Our method takes the 2D coordinate concatenated with the detection confidence as the input $\mathbf{X} \in \mathbb{R}^{T\times J \times 3}$, where $T$ and $J$ represent the input frame length and the joint coordinate number, respectively. The input is then mapped to a d-dimension feature $\mathbf{H}_{joint} \in \mathbb{R}^{ T \times J \times d}$. Simultaneously, the same input $\mathbf{X}$ also proceeds into the anatomy extraction and kinematic fusion branch to tokenize $\mathbf{H}_{bone} \in \mathbb{R}^{T \times B \times d}$ as well as $\mathbf{H}_{limb} \in \mathbb{R}^{T \times L \times d}$, where $B$ and $L$ symbolize the number of extracted bones and composed limbs, respectively. Each feature is added with a corresponding positional encoding thereafter. It is worth noting that we define $B$ and $L$ as the same value as $J$ in our practice.

The feature is further encoded with an $N$ layer multistream spatio-temporal transformer, and thereafter a regression head is applied to estimate the 3D Pose sequence output $\hat{\mathbf{P}} \in \mathbb{R}^{T \times J \times 3}$. Similar to MotionAGFormer \cite{mehraban2024motionagformer}, we adopt joint position loss $L_{pose}$ and velocity loss $L_{v}$, which are defined as
\begin{equation}
    \begin{split}
        L_{pose} &= \Sigma^{T}_{t=1} \Sigma^{J}_{j=1} \Vert \hat{\mathbf{P}}_{t,j} - \mathbf{P}_{t,j} \Vert, \\
        L_{v} &= \Sigma^{T}_{t=2} \Sigma^{J}_{j=1} \Vert \mathbf{\Delta \hat{P}}_{t,j} - \mathbf{\Delta P}_{t,j} \Vert,
    \end{split}
\end{equation}
where $\mathbf{P}$ symbolizes the Ground Truth (GT) and $\mathbf{\Delta \hat{P}_t} = \mathbf{\hat{P}_t} - \mathbf{\hat{P}_{t-1}}$, $\mathbf{\Delta P_t} = \mathbf{P_t} - \mathbf{P_{t-1}}$. The overall loss is thus defined as
\begin{equation}
    L = L_{pose} + \lambda_{v}L_{v},
\end{equation}
where the constant coefficient $\lambda_{v}$ is incorporated as a balancing factor for the smoothness of estimated motion sequences.

\subsection{Anatomy Feature Construction}

We additionally construct anatomy features through extraction and combination operations to enhance the kinematic focus of sports motions. These features are further tokenized and proceed along with the coordinate joint input. 

\noindent\textbf{Anatomy Bone Extraction (BoneExt).} As studied by Chen \textit{et al.} \cite{chen2021anatomy}, flat 2D coordinate skeleton input can be reconstructed given the length and direction of each coordinate-oriented vector. We separate the 2D coordinate part of input $\mathbf{X}$, and represent the human pose in a frame as $[\boldsymbol{z_1}...\boldsymbol{z_J}]^T \in \mathbb{R}^{J\times2}$. We treat each coordinate joint as a node in a graph-like data structure and connect two adjacent nodes to form directed edges according to a predefined skeleton direction. To be specific, we first place the root node in the pelvis and label it $0$. Then, for each joint $\boldsymbol{z_i} \in \mathbb{R}^2$, we select the closest joint node to the root as its descendant. After that, the 2D pose is capable of being decomposed into bone directions $\boldsymbol{Dir^\prime}=[\boldsymbol{dir_1}...\boldsymbol{dir_{J-1}}]^T \in \mathbb{R}^{(J-1) \times 2}$ and bone length $Len^\prime =[len_1...len_{J-1}]^T \in \mathbb{R}^{(J-1)\times 1}$ with the following equations:
\begin{equation}
    \begin{split}
        len_i &= \Vert \boldsymbol{x_i} - \boldsymbol{x_{descendant(i)}} \Vert_{2}, \\
        \boldsymbol{dir_i} &= \frac{\boldsymbol{x_i} - \boldsymbol{x_{descendant(i)}}}{len_i},
    \end{split}
\end{equation}

Subsequently, we define ($J-1$) directed edges and regard them as bones accordingly, where each bone feature contains: 1) the horizontal component value of the unit bone vector, 2) the vertical counterpart, and 3) the length of its edge. In addition, we average all defined bones to manufacture an ultra-bone with length $len_{avg} = Avg(Len^\prime) \in \mathbb{R}^{1 \times 1}$ and direction $\boldsymbol{dir_{avg}} = Avg(\boldsymbol{Dir^\prime}) \in \mathbb{R}^{1 \times 2}$. In this account, we construct the additional input $\mathbf{X_{bone}^{frame}} \in \mathbb{R}^{ J \times 3}$ for each frame and concatenate bones through frames into $\mathbf{X_{bone}} \in \mathbb{R}^{T \times J \times 3}$ with: 
\begin{equation}
    \begin{split}
        Len &= \mathcal{C}(Len^\prime, len_{avg}) \in \mathbb{R}^{J \times 1} \\
        \boldsymbol{Dir} &= \mathcal{C}(\boldsymbol{Dir^\prime},\boldsymbol{dir_{avg}}) \in \mathbb{R}^{J\times 2} \\
        \mathbf{X_{bone}^{frame}} &= \mathcal{C}(Len, \boldsymbol{Dir}) \in \mathbb{R}^{J \times 3}
    \end{split}
\end{equation}
where $\mathcal{C}$ denotes the concatenation operation, and we linearize them correspondingly into a d-dimension bone feature vector $\mathbf{H_{bone}}\in \mathbb{R}^{T \times J \times d}$.

\noindent\textbf{Kinematic Limb-wise Fusion (LimbFus).} Apart from individual bones, limb-wise mutual interactions within the human motion pattern are also taken into our vision, which is essential for grasping sports scenario context. Taking into account that our bones are described as $\boldsymbol{Dir}$ and $Len$ in the previous step, the naive vector vector-sum approach (e.g., chaining arm part or leg part vectors) fails to preserve intermediate turning point geometry information, which is vital for accurately interporating kinematic variations in limb movements, such as the bending of an elbow or knee. For the purpose of a more realistic modeling of kinematic skeleton motion, we propose to compose limbs with a dimension-wise separated encoding strategy. 

In our practice, we manually define $J$ types of limbs, eacy type of limb is a combination of $m_i, (i=1...J)$ numbers of separate bones which are grouped as each $M_{bone}$ in the subset of $M_{bones}=\{M^{1}_{bone},M^{2}_{bone},...,M^{m}_{bone} \} \in \mathbb{R}^{m_i\times 3}$. We compose our aggregated bone components to construct various limbs, employing biologically accurate limbs (arms, legs) to emphasize a localized focus on sports motion dynamics. Inspired by HDFormer \cite{chen2023hdformer}, where a series of joints and bones are selected and consciously connected with a shortest path algorithm, we likewise intentionally manufacture connections which are not present in the biological skeleton structure. In particular, we fuse some bones in imaginary hyper-limb connections (e.g., left shoulder to right hip, right arm to left leg, etc.) to study the holistic kinematic motion harmony in sports actions. Subsequently, we apply the composers $P_{limb} = [\mathcal{P}_1...\mathcal{P}_J]$ to the corresponding limbs as $\mathcal{L}_i = \mathcal{{P}}_i(M^{i}_{bone})\in \mathbb{R}^{3}$ to encode the fused limb tokens $\mathbf{X_{limb}^{frame}} = [\mathcal{L}_1...\mathcal{L}_J]^T \in \mathbb{R}^{J \times 3}$. Specifically, the bone feature for $M^{i}_{bone}$ was extracted in previous BoneExt step, which includes $\{r_{dirx}^{i}, r_{diry}^{i},r_{len}^{i}\} \in \mathbb{R}^{3 \times m_i}$. Lastly, the composers $P_{limb}$ can be defined as follows:
\begin{small}
\begin{equation}
        \mathcal{{P}}_i(r_{dirx}^{i}, r_{diry}^{i},r_{len}^{i} ) = \mathcal{C}(\mathcal{M}_X^{i}(r_{dirx}^{i}), \mathcal{M}_Y^{i}(r_{diry}^{i}), \mathcal{M}_L^{i}(r_{len}^{i})),
\end{equation} 
\end{small}
where $\mathcal{M}$ denotes a multilayer perceptron (MLP) with hidden dimension $hid$ (See Section~\ref{exp:ablation}). By such means, we construct the fused limb $\mathbf{X_{limb}} \in \mathbb{R}^{T \times J \times 3}$ and, analogously, the linearized feature $\mathbf{H_{limb}} \in \mathbb{R}^{T \times J \times d}$.

\subsection{Anatomy Feature-Enhanced Transformer}
\label{subsec:transformer}

\textbf{Preliminary.} In line with MotionAGFormer \cite{mehraban2024motionagformer}, we adopt the Metaformer \cite{yu2022metaformer} architecture to build our core encoder. Specifically, the original attention mechanism inside the transformer can be substituted with any module capable of blending information among tokens, namely the token mixer, and is formulated as
\begin{equation}
    Y = \mathrm{TokenMixer}(\mathrm{Norm}(X)) + X.
\end{equation}
Our proposed method parallelly encodes the anatomy stream and the joint stream, each contributing uniquely to comprehending the sports motion. 

\begin{table*}[htbp]
\centering
\begin{tabular}{lcccccccc} 
\hline
\multirow{2}{*}{Method}                                 & \multirow{2}{*}{Param} & \multirow{2}{*}{$N$} & \multicolumn{3}{c}{SportsPose (DET)}                                                                                                                                                                                                                                                                                                               & \multicolumn{2}{c}{SportsPose (GT)}                                                                                                                                                                                                                                                                                                     & \multirow{2}{*}{Publication}                  \\ 
\cdashline{4-8}
                                                        &                        &                    & Detector & MPJPE \textcolor[rgb]{0,0.725,0.949}{↓}                                                                                                                            & P-MPJPE \textcolor[rgb]{0,0.725,0.949}{↓}                                                                                                                          & MPJPE \textcolor[rgb]{0,0.725,0.949}{↓}                                                                                                                            & P-MPJPE \textcolor[rgb]{0,0.725,0.949}{↓}                                                                                                                          &                                               \\ 
\hline
MixSTE \cite{zhang2022mixste}                                                 & 33.6 M                 & 27                 & HRNet    & 65.2                                                                                                                                                               & 49.1                                                                                                                                                               & 33.0                                                                                                                                                               & 29.6                                                                                                                                                               & \textcolor[rgb]{0.502,0.502,0.502}{CVPR'22}   \\
STCFormer \cite{tang20233d}                                             & 4.7 M                  & 27                 & HRNet    & 64.9                                                                                                                                                               & 50.1                                                                                                                                                               & 36.4                                                                                                                                                               & 32.7                                                                                                                                                               & \textcolor[rgb]{0.502,0.502,0.502}{CVPR'23}   \\
STCFormer-L \cite{tang20233d}                                            & 18.9 M                 & 27                 & HRNet    & 64.3                                                                                                                                                               & 49.5                                                                                                                                                               & 36.0                                                                                                                                                               & 32.2                                                                                                                                                               & \textcolor[rgb]{0.502,0.502,0.502}{CVPR'23}   \\
MotionBERT \cite{zhu2023motionbert}                                             & 42.3 M                 & 27                 & HRNet    & 61.4                                                                                                                                                               & 46.8                                                                                                                                                               & 33.3                                                                                                                                                               & 30.2                                                                                                                                                               & \textcolor[rgb]{0.502,0.502,0.502}{ICCV'23}   \\
MotionAGFormer-B \cite{mehraban2024motionagformer}                                     & 11.7 M                  & 27                 & HRNet    & 60.6                                                                                                                                                               & 46.3                                                                                                                                                               & 32.6                                                                                                                                                               & 29.4                                                                                                                                                               & \textcolor[rgb]{0.502,0.502,0.502}{CVPR'23}   \\
MotionAGFormer-L \cite{mehraban2024motionagformer}                                       & 18.9 M                 & 27                 & HRNet    & \uline{59.5}                                                                                                                                                       & 47.1                                                                                                                                                               & \uline{31.3}                                                                                                                                                       & 29.3                                                                                                                                                               & \textcolor[rgb]{0.502,0.502,0.502}{CVPR'23}   \\
HDFormer \cite{chen2023hdformer}                                               & 13.4 M                 & 27                 & HRNet    & 67.7                                                                                                                                                               & 52.6                                                                                                                                                               & 35.4                                                                                                                                                               & 31.9                                                                                                                                                               & \textcolor[rgb]{0.502,0.502,0.502}{IJCAI'23}  \\
D3DP \cite{shan2023diffusion}                                                    & 34.7 M                 & 27                 & HRNet    & 60.0                                                                                                                                                               & \uline{45.1}                                                                                                                                                       & 31.9                                                                                                                                                               & \uline{28.0}                                                                                                                                                       & \textcolor[rgb]{0.502,0.502,0.502}{CVPR'24}   \\
KTPFormer \cite{peng2024ktpformer}                                              & 34.7 M                 & 27                 & HRNet    & 61.3                                                                                                                                                               & 46.7                                                                                                                                                               & 35.6                                                                                                                                                               & 31.1                                                                                                                                                               & \textcolor[rgb]{0.502,0.502,0.502}{CVPR'24}   \\ 
\hline
\rowcolor[rgb]{0.949,0.949,0.949} \textbf{KASportsFormer} (Ours) & 29.3 M                 & 27                 & HRNet    & \begin{tabular}[c]{@{}>{\cellcolor[rgb]{0.949,0.949,0.949}}c@{}}\textbf{58.0}\textcolor[rgb]{0,0.725,0.949}{}\\\textcolor[rgb]{0,0.725,0.949}{(-1.5)}\end{tabular} & \begin{tabular}[c]{@{}>{\cellcolor[rgb]{0.949,0.949,0.949}}c@{}}\textbf{44.3}\textcolor[rgb]{0,0.725,0.949}{}\\\textcolor[rgb]{0,0.725,0.949}{(-0.8)}\end{tabular} & \begin{tabular}[c]{@{}>{\cellcolor[rgb]{0.949,0.949,0.949}}c@{}}\textbf{30.9}\textcolor[rgb]{0,0.725,0.949}{}\\\textcolor[rgb]{0,0.725,0.949}{(-0.4)}\end{tabular} & \begin{tabular}[c]{@{}>{\cellcolor[rgb]{0.949,0.949,0.949}}c@{}}\textbf{27.9}\textcolor[rgb]{0,0.725,0.949}{}\\\textcolor[rgb]{0,0.725,0.949}{(-0.1)}\end{tabular} & \multicolumn{1}{l}{}                         \\
\hline
\end{tabular}
\caption{\textbf{Quantitative comparison with 3D HPE methods on the SportsPose \cite{ingwersen2023sportspose} dataset}. $N\ (=27)$: the number of input frames. DET (HRNet): using HRnet \cite{sun2019deep} detector estimated 2D keypoints as the inputs. GT: using the projected ground truth 2D keypoints as the inputs. The best and second-best results are highlighted in \textbf{bold} and \uline{underlined} formats.}
\label{tab:sportspose}
\end{table*}

\noindent\textbf{Anatomy Mixer.} 
Within athletic sports actions, we observe that bone length generally remains in a steady variation range through a spin or jump. Hence, the connections between joint coordinates, which we constructed as anatomy structure in the previous step, contain a more stabilized description guidance for analyzing fast-moving and complex sports scenarios. Hereby, we design an anatomy structure involved token mixer to explore the implicit kinematic feature through sports motion sequences. Our anatomy mixer aims at taking advantage of the bone and limb features to reinforce kinematic comprehension of human skeleton pattern, which plays an effective role in balancing insufficient temporal context caused by short video sources.

Specifically, we treat the bone and limb as different modalities of token, since both the single bone kinematic pattern and the limb-wise interaction should be taken into consideration simultaneously to reconstruct a more natural sports pose. Accordingly, we employ the Multi-Head Cross-Attention (MHCA) mechanism \cite{vaswani2017attention} to integrate potential kinematic features in sports movements. We emphasize spatial anatomy correlations as well as temporal consistencies with a Spatial MHCA (S-MHCA) $\mathcal{A}_C^S$ followed by a Temporal MHCA (T-MHCA) $\mathcal{A}_C^T$ structure. We initialize the spatial query matrix $Q_C^{S} = \mathbf{H_{bone}}W_{CQ}^{S}$ with $W$ denoting the weight matrix, while the spatial key and value matrix are always calculated from the limb feature as $K_C^{S} = \mathbf{H_{limb}}W_{CK}^{S}$ and $V_C^{S} = \mathbf{H_{limb}}W_{CV}^{S}$ seeing that these limbs we constructed are composed of multiple bones and with potentially more kinematicly encoded information, thus we select the $\mathbf{H_{limb}}$ to project the value vectors.

Subsequently, we transpose the spatial output to the temporal input $\mathbf{\bar{H}_{bone}^{ i}}$ and $\mathbf{\bar{H}_{limb}^{i}}$, where $\mathbf{\bar{H}_{bone}} = (\mathbf{H_{bone}})^T$ and $\mathbf{\bar{H}_{limb}} = (\mathbf{H_{limb}})^T$, respectively. We also apply the projection to produce $Q_C^{Ti} = \mathbf{\bar{H}_{bone}}W_{CQ}^{Ti}$, $K_C^{Ti} = \mathbf{\bar{H}_{limb}}W_{CK}^{Ti}$ and $V_C^{Ti} = \mathbf{\bar{H}_{limb}}W_{CV}^{Ti}$. The calculation can be annotated as: 
\begin{equation}
    \begin{split}
        \mathcal{A}_C^S = \mathcal{C}(head_{C}^{S1}...head_{C}^{Sh}){W_{CO}^S}, \\
        head_C^{Si} = \mathrm{softmax}(\frac{Q_C^{Si}(K_C^{Si})^T}{\sqrt{d_k}})V_C^{Si}, \\
        \mathcal{A}_C^T = \mathcal{C}(head_{C}^{T1}...head_{C}^{Th}){W_{CO}^T}, \\
        head_C^{Ti} = \mathrm{softmax}(\frac{Q_C^{Ti}(K_C^{Ti})^T}{\sqrt{d_k}})V_C^{Ti}, \\
    \end{split}
    \label{eq:mhca}
\end{equation}
where $(i=1...h)$ and $h$ denotes the numbers of the attention head. $d_k$ is the dimension of the corresponding key $K$ matrix. The attention results are further proceeded into an MLP module, followed by the residual connection and Layer Normalization.

\noindent\textbf{Joint Mixer.} We preserve the joint coordinates in our proposed KASportsFormer for their significance in fine-grained positional locating for reconstructed sports poses. Referring to MotionAGFormer \cite{mehraban2024motionagformer}, we encode the joint stream with the Multi-Head Self-Attention (MHSA) mechanism. To ensure a thorough exploration of local and global kinematic relations, the joint feature $\mathbf{H_{joint}}$ is first mixed with a Spatial MHSA (S-MHSA) $\mathcal{A}_{S}$ and then transposed to forward into a Temporal MHSA (T-MHSA) $\mathcal{A}_{T}$. The overall process is similar to the calculation (\ref{eq:mhca}).

Parallelly, the Joint Mixer also takes advantage of Graph Convolutional Network (GCN) structure as an incorporated module, which provides an amplified focus on local joint space and a sliding window strategy for crucial frame emphasis. This equips the joint mixer with a more effective balance in local and global kinematic cues of sports movements. The Spatial GCN $\mathcal{G_S}$ constructs the self-connected adjacency matrix $\widetilde{A} = A + I$, where $I$ is the identity matrix. We adopt the matrix $\widetilde{D}$, in which $\widetilde{D}_{ii} = \Sigma_{j}\widetilde{A}_{jj}$ is the matrix expression of the token mixture process.
\begin{equation}
    \mathcal{G} = \sigma(\mathbf{H_{joint}} + \mathcal{N}(\widetilde{D}^{-\frac{1}{2}}\widetilde{A}\widetilde{D}^{-\frac{1}{2}}\mathbf{H_{joint}}W_1 + \mathbf{H_{joint}}W_2)), 
\end{equation}
where $\sigma$ and $\mathcal{N}$ denote the ReLu and normalization, respectively. The temporal counterpart $\mathcal{G_T}$ implements local highlightings along frame sequences with the strategy of connecting $K$ highest reactions nodes through frame similities $Sim(\mathbf{t_i}, \mathbf{t_j}) = (\mathbf{\bar{H}_{joint}^{t_i}})^T\mathbf{\bar{H}_{joint}^{t_j}}$ to construct adjacency matrix $\widetilde{A}$.

\noindent\textbf{Token Blending.} 

We assign the each stream output of our anatomy feature-enhanced transformer as $\{\mathbf{H_{\mathcal{A_C}}^{i}},\mathbf{H_{\mathcal{A_S}}^{i}}, \mathbf{H_{\mathcal{G}}^{i}}\}, (i=1...N)$ corresponding to MHCA, MHSA, and GCN mixer stream, where $i$ symbolizes the current layer index. Purposing for constructing the feature-wise communication, we adopt a adaptive blending strategy at each layer end as MotionBERT \cite{zhu2023motionbert}, to aggregate each perspective of sports motion perception, which can be defined as: 
\begin{equation}
\label{eq:softmax1}
    \mathbf{H^i} = \alpha_{\mathcal{A_C}}^i \odot \mathbf{H_{\mathcal{A_C}}^{i-1}} + \alpha_{\mathcal{A_S}}^i \odot \mathbf{H_{\mathcal{A_S}}^{i-1}} + \alpha_{\mathcal{G}}^i \odot \mathbf{H_{\mathcal{G}}^{i-1}},
\end{equation}
where $\alpha$ denotes the knowledge factor for each mixer stream, which is calculated with a softmax on each mixer output, and $\odot$ symbolizes element-wise multiplication. The calculation can be expressed as:
\begin{equation}
\label{eq:softmax2}
\alpha_{\mathcal{A_C}}^i,\alpha_{\mathcal{A_S}}^i,\alpha_{\mathcal{G}}^i = \mathrm{softmax}(W \cdot \mathcal{C}(\mathbf{H_{\mathcal{A_C}}^{i-1}}, \mathbf{H_{\mathcal{A_S}}^{i-1}}, \mathbf{H_{\mathcal{G}}^{i-1}})),
\end{equation}
with $W$ representing a learnable weight matrix.

Noting worthily, we utilize the blended feature $\mathbf{H^i}$ to tokenize joint and bone feature input for the next transformer structure layer $i + 1$, as $\mathbf{H_{joint}^i} = \mathbf{H^i}$ and $\mathbf{H_{bone}^i} = \mathbf{H^i}$. While on the contrary, limb feature $\mathbf{H_{limb}}$ is always encoded with the same input $\mathbf{X_{limb}}$, owing to our motivation for constructing a kinematic context for skeleton comprehension. Lastly, we apply an MLP mapping module, which is denoted as regression head, on the linearized final layer output to regress the 3D sports pose sequence $\mathbf{\hat{P}}$ as our estimation results.

\section{Experiments}
\label{sec:experiments}

We evaluate the performance of our proposed KASportsFormer on two representative and relatively recently released sports-centered 3D human pose estimation datasets, i.e, SportsPose \cite{ingwersen2023sportspose} and WorldPose \cite{jiang2024worldpose}.

\begin{table*}[htbp]
\centering
\begin{tabular}{lccccc;{1pt/1pt}ccccc} 
\hline
\multirow{2}{*}{Method / Action}                        & \multicolumn{5}{c}{SportsPose (DET) MPJPE \textcolor[rgb]{0,0.722,0.949}{↓}}                                                                                                                                                                                                                                                                                                                                                                                                                                                                                                                                                                                              & \multicolumn{5}{c}{SportsPose (GT) MPJPE \textcolor[rgb]{0,0.722,0.949}{↓}}                                                                                                                                                                                                                                                                                                                                                                                                                                                                                                                                                                                                             \\ 
\cdashline{2-11}
                                                        & Throw                                                                                                                              & Soccer                                                                                            & Tennis                                                                                                                             & Jump                                                                                                                               & \multicolumn{1}{c}{Volley}                                                                                                                             & Throw                                                                                                                              & Soccer                                                                                                                             & Tennis                                                                                                                             & Jump                                                                                                                               & Volley                                                                                                                              \\ 
\hline
MixSTE \cite{zhang2022mixste}                                                  & 68.5                                                                                                                               & 60.6                                                                                              & 60.9                                                                                                                               & 78.1                                                                                                                               & 58.2                                                                                                                                                   & 38.9                                                                                                                               & 32.2                                                                                                                               & 30.7                                                                                                                               & 30.7                                                                                                                               & 32.8                                                                                                                                \\
STCFormer \cite{tang20233d}                                              & 68.9                                                                                                                               & 59.7                                                                                              & 60.5                                                                                                                               & 77.6                                                                                                                               & 58.1                                                                                                                                                   & 42.4                                                                                                                               & 35.4                                                                                                                               & 32.9                                                                                                                               & 36.5                                                                                                                               & 35.2                                                                                                                                \\
STCFormer-L \cite{tang20233d}                                            & 67.6                                                                                                                               & 60.3                                                                                              & 58.7                                                                                                                               & 77.8                                                                                                                               & 57.5                                                                                                                                                   & 42.5                                                                                                                               & 34.4                                                                                                                               & 32.5                                                                                                                               & 35.6                                                                                                                               & 35.2                                                                                                                                \\
MotionBERT \cite{zhu2023motionbert}                                             & 64.7                                                                                                                               & 58.4                                                                                              & 56.8                                                                                                                               & 71.2                                                                                                                               & 56.1                                                                                                                                                   & 40.0                                                                                                                               & 32.0                                                                                                                               & 31.2                                                                                                                               & 30.6                                                                                                                               & 33.0                                                                                                                                \\
MotionAGFormer-B \cite{mehraban2024motionagformer}                                      & 67.5                                                                                                                               & \textbf{56.3}                                                                                              & \uline{56.2}                                                                                                                               & \uline{68.3}                                                                                                                               & 55.0                                                                                                                                                   & 37.3                                                                                                                               & 32.1                                                                                                                               & 30.6                                                                                                                               & 31.0                                                                                                                               & 32.1                                                                                                                                \\
MotionAGFormer-L \cite{mehraban2024motionagformer}                                       & 63.1                                                                                                                               & 56.4                                                                                     & 56.3                                                                                                                       & 69.4                                                                                                                               & \uline{52.7}                                                                                                                                           & \uline{35.9}                                                                                                                       & 30.9                                                                                                                               & 30.0                                                                                                                               & \uline{29.1}                                                                                                                       & \uline{30.9}                                                                                                                        \\
HDFormer \cite{chen2023hdformer}                                               & 73.1                                                                                                                               & 61.9                                                                                              & 62.4                                                                                                                               & 81.4                                                                                                                               & 59.6                                                                                                                                                   & 41.4                                                                                                                               & 34.4                                                                                                                               & 32.6                                                                                                                               & 34.5                                                                                                                               & 34.5                                                                                                                                \\
D3DP \cite{shan2023diffusion}                                                   & \uline{62.1}                                                                                                                       & 57.4                                                                                              & 56.4                                                                                                                               & 69.2                                                                                                                       & 55.2                                                                                                                                                   & 37.4                                                                                                                               & \uline{30.8}                                                                                                                       & \uline{29.6}                                                                                                                       & 30.0                                                                                                                               & 31.9                                                                                                                                \\
KTPFormer \cite{peng2024ktpformer}                                              & 65.2                                                                                                                               & 58.3                                                                                              & 57.7                                                                                                                               & 70.8                                                                                                                               & 54.7                                                                                                                                                   & 42.0                                                                                                                               & 34.8                                                                                                                               & 32.8                                                                                                                               & 33.7                                                                                                                               & 34.8                                                                                                                                \\ 
\hline
\rowcolor[rgb]{0.949,0.949,0.949} \textbf{KASportsFormer} (Ours) & \begin{tabular}[c]{@{}>{\cellcolor[rgb]{0.949,0.949,0.949}}c@{}}\textbf{60.6}\\\textcolor[rgb]{0,0.722,0.949}{(-1.5)}\end{tabular} & \begin{tabular}[c]{@{}>{\cellcolor[rgb]{0.949,0.949,0.949}}c@{}}\uline{57.0}\\(+0.7)\end{tabular} & \begin{tabular}[c]{@{}>{\cellcolor[rgb]{0.949,0.949,0.949}}c@{}}\textbf{54.3}\\\textcolor[rgb]{0,0.722,0.949}{(-1.9)}\end{tabular} & \begin{tabular}[c]{@{}>{\cellcolor[rgb]{0.949,0.949,0.949}}c@{}}\textbf{66.6}\\\textcolor[rgb]{0,0.722,0.949}{(-1.7)}\end{tabular} & \multicolumn{1}{c}{\begin{tabular}[c]{@{}>{\cellcolor[rgb]{0.949,0.949,0.949}}c@{}}\textbf{51.8}\\\textcolor[rgb]{0,0.722,0.949}{(-0.9)}\end{tabular}} & \begin{tabular}[c]{@{}>{\cellcolor[rgb]{0.949,0.949,0.949}}c@{}}\textbf{35.6}\\\textcolor[rgb]{0,0.722,0.949}{(-0.3)}\end{tabular} & \begin{tabular}[c]{@{}>{\cellcolor[rgb]{0.949,0.949,0.949}}c@{}}\textbf{30.1}\\\textcolor[rgb]{0,0.722,0.949}{(-0.7)}\end{tabular} & \begin{tabular}[c]{@{}>{\cellcolor[rgb]{0.949,0.949,0.949}}c@{}}\textbf{29.3}\\\textcolor[rgb]{0,0.722,0.949}{(-0.3)}\end{tabular} & \begin{tabular}[c]{@{}>{\cellcolor[rgb]{0.949,0.949,0.949}}c@{}}\textbf{28.8}\\\textcolor[rgb]{0,0.722,0.949}{(-0.3)}\end{tabular} & \begin{tabular}[c]{@{}>{\cellcolor[rgb]{0.949,0.949,0.949}}c@{}}\textbf{30.7}\\\textcolor[rgb]{0,0.722,0.949}{(-0.2)}\end{tabular}  \\
\hline
\end{tabular}
\caption{\textbf{Action based quantitative comparison with 3D HPE methods on the SportsPose \cite{ingwersen2023sportspose} dataset using both detected and ground truth projected keypoints as the inputs.} Throw, Soccer, Tennis, Jump, and Volley refers to 5 sports action. The best and second-best results are highlighted in \textbf{bold} and \uline{underlined} formats.}
\label{tab:sportspose_action}
\end{table*}


\begin{table*}[htbp]
\centering
\begin{tabular}{lcccccccc} 
\hline
\multirow{2}{*}{Method}                                          & \multirow{2}{*}{Param} & \multirow{2}{*}{$N$} & \multicolumn{3}{c}{WorldPose (DET)}                                                                                                                                                                                                                                                & \multicolumn{2}{c}{WorldPose (GT)}                                                                                                                                                                                                                                    & \multirow{2}{*}{Publication}                  \\ 
\cdashline{4-8}
                                                                 &                        &                    & Detector & MPJPE \textcolor[rgb]{0,0.722,0.949}{↓}                                                                                            & P-MPJPE \textcolor[rgb]{0,0.722,0.949}{↓}                                                                                          & MPJPE \textcolor[rgb]{0,0.722,0.949}{↓}                                                                                           & P-MPJPE \textcolor[rgb]{0,0.722,0.949}{↓}                                                                                         &                                               \\ 
\hline
MixSTE \cite{zhang2022mixste}                                                          & 33.6 M                 & 27                 & HRNet    & 52.8                                                                                                                               & 30.7                                                                                                                               & 25.3                                                                                                                              & 11.0                                                                                                                              & \textcolor[rgb]{0.502,0.502,0.502}{CVPR'22}   \\
STCFormer \cite{tang20233d}                                                       & 4.7 M                  & 27                 & HRNet    & 44.1                                                                                                                               & 28.8                                                                                                                               & 14.2                                                                                                                              & 10.7                                                                                                                              & \textcolor[rgb]{0.502,0.502,0.502}{CVPR'23}   \\
STCFormer-L \cite{tang20233d}                                                     & 18.9 M                 & 27                 & HRNet    & 42.0                                                                                                                               & 27.3                                                                                                                               & 13.3                                                                                                                              & 10.3                                                                                                                              & \textcolor[rgb]{0.502,0.502,0.502}{CVPR'23}   \\
MotionBERT \cite{zhu2023motionbert}                                                      & 42.3 M                 & 27                 & HRNet    & 36.3                                                                                                                               & 23.3                                                                                                                               & \uline{9.3}                                                                                                                       & \uline{6.8}                                                                                                                       & \textcolor[rgb]{0.502,0.502,0.502}{ICCV'23}   \\
MotionAGFormer-B \cite{mehraban2024motionagformer}                                                & 11.7 M                & 27                 & HRNet    & \uline{35.6}                                                                                                                       & \uline{22.9}                                                                                                                       & 11.4                                                                                                                              & 8.1                                                                                                                               & \textcolor[rgb]{0.502,0.502,0.502}{CVPR'23}   \\
MotionAGFormer-L \cite{mehraban2024motionagformer}                                                & 18.9 M                 & 27                 & HRNet    & 35.9                                                                                                                               & 23.1                                                                                                                               & 11.1                                                                                                                              & 8.3                                                                                                                               & \textcolor[rgb]{0.502,0.502,0.502}{CVPR'23}   \\
HDFormer \cite{chen2023hdformer}                                                        & 13.4 M                 & 27                 & HRNet    & 52.1                                                                                                                               & 35.1                                                                                                                               & 18.8                                                                                                                              & 14.9                                                                                                                              & \textcolor[rgb]{0.502,0.502,0.502}{IJCAI'23}  \\
D3DP \cite{shan2023diffusion}                                                            & 34.7 M                 & 27                 & HRNet    & 47.9                                                                                                                               & 24.9                                                                                                                               & 18.8                                                                                                                              & 8.9                                                                                                                               & \textcolor[rgb]{0.502,0.502,0.502}{CVPR'24}   \\
KTPFormer \cite{peng2024ktpformer}                                                       & 34.7 M                 & 27                 & HRNet    & 36.7                                                                                                                               & 23.7                                                                                                                               & 13.4                                                                                                                              & 8.8                                                                                                                               & \textcolor[rgb]{0.502,0.502,0.502}{CVPR'24}   \\ 
\hline
\rowcolor[rgb]{0.949,0.949,0.949} \textbf{KASportsFormer} (Ours) & 29.3 M                 & 27                 & HRNet    & \begin{tabular}[c]{@{}>{\cellcolor[rgb]{0.949,0.949,0.949}}c@{}}\textbf{34.2}\\\textcolor[rgb]{0,0.722,0.949}{(-1.4)}\end{tabular} & \begin{tabular}[c]{@{}>{\cellcolor[rgb]{0.949,0.949,0.949}}c@{}}\textbf{22.0}\\\textcolor[rgb]{0,0.722,0.949}{(-0.9)}\end{tabular} & \begin{tabular}[c]{@{}>{\cellcolor[rgb]{0.949,0.949,0.949}}c@{}}\textbf{8.5}\\\textcolor[rgb]{0,0.722,0.949}{(-0.8)}\end{tabular} & \begin{tabular}[c]{@{}>{\cellcolor[rgb]{0.949,0.949,0.949}}c@{}}\textbf{6.2}\\\textcolor[rgb]{0,0.722,0.949}{(-0.4)}\end{tabular} & \multicolumn{1}{l}{}                          \\
\hline
\end{tabular}
\caption{\textbf{Qualitative comparison with the 3D HPE methods on the WorldPose \cite{jiang2024worldpose} dataset.} $N$: the number o input frames is fixed to 27. DET(HRNet): using HRNet \cite{sun2019deep} detector estimated 2D keypoints as the inputs. GT: using projected ground truth 2D keypoints as the inputs. The best and second-best results are highlighted in \textbf{bold} and \uline{underlined} formats.}
\label{tab:worldpose}
\end{table*}

\subsection{Datasets and Metrics}

\noindent\textbf{SportsPose} \cite{ingwersen2023sportspose} is a recently published 3D human pose estimation dataset, which contains a large-scale collection of 3D human joint keypoint annotations focusing on sports activities.  This dataset consists of 1.5 million video 
frames recorded from 7 cameras located at different angles, containing 24 subjects performing 5 different sports activities (e.g., Soccer, Volleyball, Tennis). Following the processing procedure of previous works \cite{pavllo20193d, zhang2022mixste, li2022mhformer} and maintaining the train and test data quantity ratio of \cite{mehraban2024motionagformer}, we randomly divide the entire dataset into train and test subsets based on subjects. In our practice, 7 subjects (i.e., S2, S4, S7, S10, S16, S21, S22) are designated for evaluation, and the remaining subjects are treated as training data. We also create the 2D estimated keypoints input using an unfinetuned 2D Pose detector, HRNet \cite{sun2019deep}, on the available video footage in order to demonstrate the model's capability against noises that may occur in real sports video applications.

Notably, the source video footage released in SportsPose \cite{ingwersen2023sportspose} is shorter than typical 3D HPE datasets \cite{h36m_pami}, resulting in insufficient training data after sampling long video clips (e.g., 243 frames). Hence, we only sample video clips with a commonly used short frame length (i.e., 27) for Detection data. To remain the consistency of SportsPose concerned experiments, we keep the sequence length to 27 in order to reveal the model's performance on short videos. We retrain all the models in our experiments with our prepared 27-frame SportsPose data.

Subsequently, we calculate the mean per joint position error (MPJPE) results to measure the average Euclidean distance in millimeters between the 3D joint position estimation and the ground truth for evaluation. We also record the procrustes MPJPE (P-MPJPE) results, which calculate MPJPE after a rigid transformation executed on estimated poses. Following these procedures and metrics, we select some recently released 3D HPE models and retrain them on our processed short video data to compare their performance with our proposed KASportsFormer.

\noindent\textbf{WorldPose} \cite{jiang2024worldpose} is a lately released sports-focused dataset, providing plentiful accurate 3D annotations based on video clips from 8 soccer games. WorldPose consists of 1.5K footage frames from games of 2022 FIFA WorldCup amounting 2.5M annotated players' SMPL \cite{SMPL2015} poses in total. The camera setup for the video footage in this dataset remains the same as that used in the soccer game broadcasting system. We include this dataset in our performance analysis with the purpose of verifying the model's applicable capability in a more realistic sports scenario.

The Worldpose dataset annotates all soccer players in the camera footage, whereas our focus is on analyzing sports poses for a single person. Intended for preparing detection data, we first project the player's 3D keypoints onto 2D footage frames using the dataset's provided camera parameters, and then we crop out each player with the bounding box of projected 2D keypoints. To maintain consistency with the SportsPose \cite{ingwersen2023sportspose} dataset, we likewise utilize the same HRNet \cite{sun2019deep} detector for 2D inputs. Due to the fact that the resolutions of cropped images are relatively lower, the 2D detector produces more anomalies than usual as a consequence. We regard this phenomenon as a unique internal feature of the WorldPose dataset itself. Additionally, we also keep the sequence length the same as the SportsPose dataset to emphasize our focus on revealing short sports video understanding capabilities. 

This dataset annotates poses based on footage clips starting from different timestamps from a total of 8 soccer games. We randomly categorize clips of each games into training and testing sets, and in practice, a total of 26 clip sources are used for performance validation, and the remaining games are utilized as training data, to maintain a similar ratio to our processed SportsPose available data clips. We also report the MPJPE and P-MPJPE scores for a comprehensive performance evaluation under a more realistic sports scenario.


\begin{table}[htbp]
\centering
\begin{tabular}{lccc} 
\hline
\multirow{2}{*}{Limb}                                & \multirow{2}{*}{Param} & \multicolumn{2}{c}{SportsPose (DET)}  \\ 
\cdashline{3-4}
                                                     &                        & MPJPE↓ & P-MPJPE↓                     \\ 
\hline
$hid=128$                                            & 29.395 M               & 59.0   & 45.1                         \\
$hid=64$                                             & 29.378 M               & 58.5   & 44.7                         \\
$hid=32$                                             & 29.369 M               & 58.7   & 45.3                         \\
$hid=8$                                              & 29.363 M               & 59.0   & 45.1                         \\ 
\hline
\rowcolor[rgb]{0.949,0.949,0.949} \textbf{Ours (16)} & 29.365 M               & \textbf{58.0} & \textbf{44.3}                \\
\hline
\end{tabular}
\caption{\textbf{Ablation study on different depth designs of LimbFus block.} $hid$: number of hidden layers inside each LimbFus block corresponding to each composed limb.}
\label{tab:limb}
\end{table}

\subsection{Implementation Details}

All of our experiments are implemented using the PyTorch \cite{paszke2017automatic} framework and executed on the NVIDIA A6000 GPU environment.
The training epoch of all experiments is set to 120, and the batch size is set to 32. In addition, we employ an early stopping strategy where the training process will be interrupted if the evaluation results are not improving for 10 epochs. We adopt an AdamW optimizer with the learning rate initially set at 5e-4 and the weight decay rate set as 0.01. We also implement a learning rate warming-up for the first 10 epochs, starting at 5e-6, and the learning rate decreases by a factor of 0.9 after the warming-up round with a 2-epoch patience PyTorch learning rate scheduler module. 

We prepare our SportsPose and WorldPose data with a 17-joint pose format in the same as Human3.6M's \cite{h36m_pami} definition instead of the dataset's provided COCO \cite{cocodataset} human skeleton format. Following the previous practice \cite{zhao2023poseformerv2, zhu2023motionbert}, we likewise apply the horizontal flipping data augmentation strategy and fix the pelvis joint on zero coordinate point for both training and testing process. Concerning our model configuration, the total number of layers is set to $N=26$ in our anatomy feature enhanced transformer structure part (section~\ref{subsec:transformer}), and the intermediate feature dimension is set to $d=128$. Inside the KASportsFormer, both the MHSA and the MHCA streams consist of a head $h=8$, and we follow MotionAGFormer \cite{mehraban2024motionagformer} to set neighbour number as $K=2$ of the customized GCN structure stream. Additionally, our LimbFus module is constructed with internal MLP networks that hold a hidden layer dimension of $hid = 16$ (see Section~\ref{exp:ablation}).

\subsection{Comparison with the State-of-the-Arts}

\noindent\textbf{SportsPose \cite{ingwersen2023sportspose}}. Table~\ref{tab:sportspose} reports the numeric comparisons between our proposed method with some recent 3D HPE methods on the SportsPose dataset. Our KASportsFormer achieved the current state-of-the-art (SOTA) performance on 27 frame video clips, using both HRNet detected and ground truth projected 2D keypoints as the input source. Compared with existing 3D HPE methods, our methods achieved a performance improvement on the previous SOTA approach, MotionAGFormer-L \cite{mehraban2024motionagformer}, with the decrease of MPJPE error rate from 59.5mm to 58.0mm. And we also achieved a 0.8mm improvement on the P-MPJPE metric compared with the diffusion-based approach D3DP \cite{shan2023diffusion} using detected 2D inputs. With regard to Ground Truth 2D input source, our method continually gained performance refinement and achieved 30.9mm and 27.9mm on the MPJPE and P-MPJPE metrics, respectively. The quantitative comparison result demonstrated that our anatomy feature involved approach improved the comprehensive capability for sports movements under short video limitations.

We further compare the detailed estimation performance on each sports activity contained in SportsPose dataset with detection input, which is shown as Table~\ref{tab:sportspose_action}. With a better understanding of kinematic consistency in sports motions, our method generally improved the numeric results to some extent, except for the detection input of soccer, and we attribute this phenomenon to our limited perception of limb acceleration in kicking movements.


\begin{table}[htbp]
\centering
\begin{tabular}{lcccc} 
\hline
\multirow{2}{*}{Method}                         & \multicolumn{3}{c}{Configuration} & \multirow{2}{*}{P1↓/P2↓}  \\ 
\cdashline{2-4}
                                                & BoneExt & LimbFus & MHCA          &                           \\ 
\hline
baseline                                        &         &         &               & 59.6/47.1                 \\
w/o bone                                        &         & \checkmark        &               & 58.6/44.4                 \\
w/o limb                                        & \checkmark        &         &               & 58.9/44.9                 \\
softmax                                        & \checkmark        & \checkmark        &               & 59.0/45.0                 \\ 
\hline
\rowcolor[rgb]{0.949,0.949,0.949} \textbf{Ours} & \checkmark        & \checkmark        & \checkmark              & \textbf{58.0/44.3}        \\
\hline
\end{tabular}
\caption{\textbf{Abalation study on different configurations of KASportsFormer with SportsPose (DET) input.} baseline: the same structure with MotionAGFormer \cite{mehraban2024motionagformer}. w/o bone: the method without any bone feature, and process the limb tokens with MHSA. w/o limb: the method without incorporation of any limb constructions. softmax: combine the bone and limb features with a softmax calculated weight. P1: MPJPE error. P2: P-MPJPE error}
\label{tab:config}
\end{table}

\noindent\textbf{WorldPose \cite{jiang2024worldpose}}. Table~\ref{tab:worldpose} reports comparisons between KASportsFormer and other 3D HPE methods on the WorldPose datasets with HRNet detected and GT projected coordinates as 2D keypoints inputs. Compared with the current SOTA method, MotionAGFormer-B \cite{mehraban2024motionagformer}, our method decreased MPJPE and P-MPJPE by 1.4mm and 0.9mm, respectively. 
Meanwhile, on the WorldPose GT data, MotionBERT \cite{zhu2023motionbert} achieved the second-best result, and we outperformed D3DP \cite{shan2023diffusion} by 0.8mm MPJPE and 0.4mm P-MPJPE, respectively. Overall, these experimental results demonstrated that utilizing the anatomy feature with our proposed KASportsFormer structure can enhance the model's understanding of human sports activities to some extent and thus improved estimation accuracy.


\begin{figure*}[htbp]
    \centering
    \includegraphics[width=1.0\linewidth]{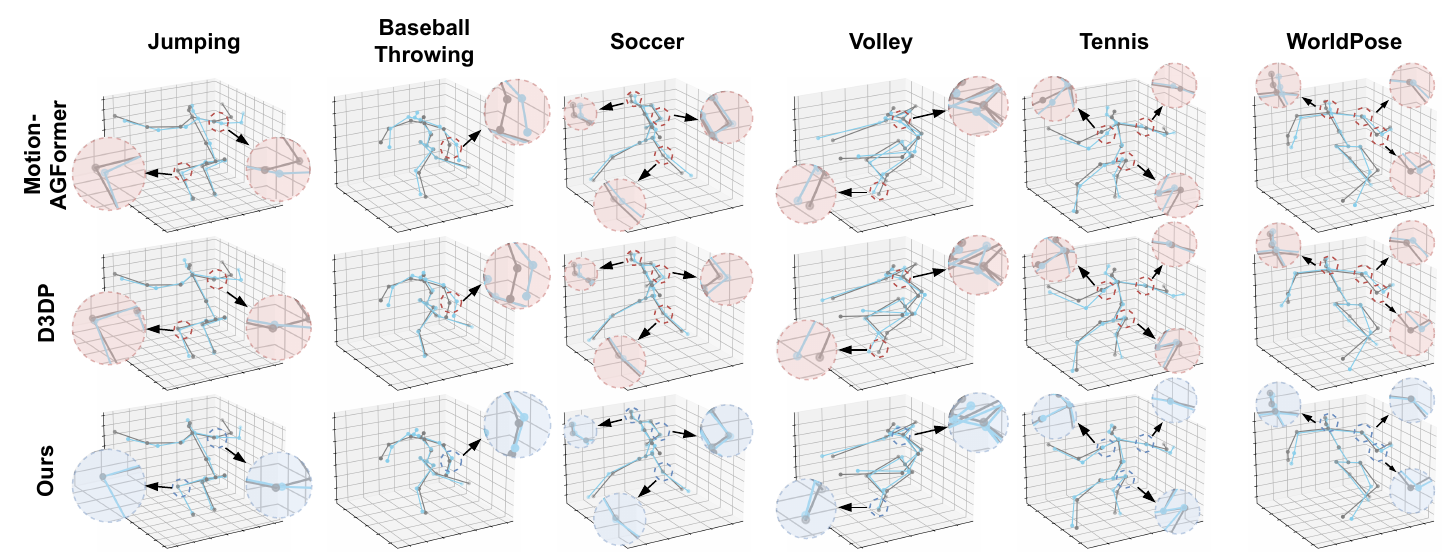}
    \caption{\textbf{Qualitative comparisons of our KASportsFormer with MotionAGFormer \cite{mehraban2024motionagformer} and D3DP \cite{shan2023diffusion} on activities from SportsPose and WorldPose.} The gray skeleton is the ground truth 3D pose. The blue skeleton represents the estimated 3D pose result. The red dashed line indicates the incorrect regions of the compared methods, and the blue dashed line represents the counterparts of our proposed method.}
    \label{fig:visualization}
\end{figure*}

\subsection{Ablation Study}
\label{exp:ablation}

We conducted a series of ablation studies to analyze the effectiveness of our KASportsFormer design for short sports video scenarios. For the simplicity and consistency of the experiment conditions, we focused on the SportsPose dataset and sampled short video clips with 27 frames. 

We first explored the parameter design of our LimbFus block, which utilizes MLPs to combine different bone parts. The parameter $hid$ denotes the number of hidden layers in each MLP block. As demonstrated in the Table~\ref{tab:limb}, the depth of the hidden layer acted as a compromise of accuracy when combining different bones. Compared with KASportsFormer, defining the hidden layer as $hid=128$, which is the same quantity as the feature dimension in our Transformer layer, caused the appearance of the underfitting phenomenon, since the heavy structure lacked an appropriate normalization strategy. Still, $hid=8$ tended to be insufficient for encoding complicated limb transformations in sports movements and also introduces overfitting. In our practice, $hid=16$ achieved a relatively stable balance, allowing the fused limb to perceive kinematic sports actions more effectively in short video situations.  

We also investigated the effect of our pipeline design, which starts with the extraction of bones, the fusion of limbs, and finally, the mutual interaction to enhance monocular sports motion reconstruction. We designed iterative versions of our KASportsFormer structure to evaluate the contributions of different blocks to sports-focused 3D HPE performance, including ``w/o bone'', ``w/o limb'', and ``softmax''. Specifically, we used the original structure of MotionAGFormer \cite{mehraban2024motionagformer} as the baseline, and we compared our design variations accordingly. ``w/o limb'' indicates that only the extracted bone feature is utilized with the MHSA mechanism, as it is the only modality. Analogously, ``w/o bone'' is designed with the only participation of limbs. ``softmax'' denotes that we substituted the MHCA module with a softmax structure, which integrates different modalities using a similar calculation as (\ref{eq:softmax1}) and (\ref{eq:softmax2}). 

As shown in Table~\ref{tab:config}, we observed that our "w/o limb" decreased both the MPJPE and the P-MPJPE error from baseline, indicating that extracting the bone feature and combining with the joint feature was an effective direction for enhancing sports motion understanding, even though the sequence context was strictly limited. In addition, "w/o bone" indicated that exploiting the composed limb feature was also effective in discovering potential consistency of complicated sports movements. As demonstrated, our KASportsFormer outperformed other versions of module structure, illustrating the effectiveness of bone extraction, limb fusion, and MHCA way of token interaction, in handling temporal context-limited sports pose estimation problems.

\subsection{Visualization}

Figure~\ref{fig:visualization} shows some of the visualized estimation results of MotionAGFormer \cite{mehraban2024motionagformer}, D3DP \cite{shan2023diffusion}, and our KASportsFormer on different SportsPose \cite{ingwersen2023sportspose} actions with detection input as well as the WorldPose \cite{jiang2024worldpose} detection input. These methods generally performed well when the body structure in the scene was relatively clear. For some athlete actions with dynamic movements (e.g., Soccer Shooting and Jumping), the 3D poses estimated by our KASportsFormer matched better with the ground truth, especially in the junction parts of different limbs, which verified the effectiveness of exploring bone continuity inside sports motion. For actions involving fast movements (e.g., throwing and volleying), our method was more accurate at tip parts, such as hands and feet, which commonly moved in an extensive range, and were thus challenging for current 3D HPE methods. In addition, our proposed method also outperformed the counterparts for some movements in the WorldPose dataset (e.g., Ball Kicking), which reflected the robustness of our method when handling more realistic sports video scenes.

\section{Conclusion}
\label{sec:conclusion}

In this work, we propose KASportsFormer, a novel anatomy feature-enhanced approach for 3D HPE, which extracts the anatomy bone structure feature from existing joint coordinates, followed by a fusion strategy for intensifying limb motion pattern perception. Our KASportsFormer block leverages these anatomy features in a multi-modality manner, which was relatively effective in capturing intricate spatial interdependencies of kinematic human motions under sports scenarios. Experimental results on two sports-based benchmarks demonstrated that KASportsFormer surpassed the current state-of-the-art methods and has a better capacity for sports motion understanding.


\bibliographystyle{ACM-Reference-Format}
\bibliography{base}


\end{document}